\documentclass{article}

\usepackage{natbib}

\usepackage[final]{neurips_2021}

\usepackage{amsmath}
\DeclareMathOperator*{\argmax}{argmax}

\usepackage{natbib}
\usepackage{graphicx}
\usepackage[utf8]{inputenc} 
\usepackage[T1]{fontenc}    
\usepackage{hyperref}       
\usepackage{url}            
\usepackage{booktabs}       
\usepackage{amsfonts}       
\usepackage{nicefrac}       
\usepackage{microtype}      
\usepackage{xcolor}         
\usepackage{lipsum}
\usepackage{pifont}
\newcommand{\cmark}{\ding{51}}%
\newcommand{\xmark}{\ding{55}}%
\usepackage{listings}
\usepackage{xcolor}
\definecolor{codegreen}{rgb}{0,0.6,0}
\definecolor{codegray}{rgb}{0.5,0.5,0.5}
\definecolor{codepurple}{rgb}{0.58,0,0.82}
\definecolor{backcolour}{rgb}{0.95,0.95,0.92}
\lstdefinestyle{mystyle}{
    backgroundcolor=\color{backcolour},   
    commentstyle=\color{codegreen},
    keywordstyle=\color{magenta},
    numberstyle=\tiny\color{codegray},
    stringstyle=\color{codepurple},
    basicstyle=\ttfamily\footnotesize,
    breakatwhitespace=false,         
    breaklines=true,                 
    captionpos=b,                    
    keepspaces=true,                 
    numbers=left,                    
    numbersep=5pt,                  
    showspaces=false,                
    showstringspaces=false,
    showtabs=false,                  
    tabsize=2
}

\def\figref#1{Fig.~\ref{#1}}

\def\eqref#1{Eq.~(\ref{#1})}
\def\secref#1{Sec.~\ref{#1}}

\def\eg{{\it e.g.}}

\def\ie{{\it i.e.}}

\def\rev#1{\textcolor{black}{#1}}

\title{ShinRL: A Library for Evaluating RL Algorithms from Theoretical and Practical Perspectives}

\author{%
  Toshinori Kitamura\thanks{Work done as an intern at OMRON SINIC X.} \\
  Nara Institute of Science and Technology\\
  Nara, Japan\\
  \texttt{kitamura.toshinori.kt6@is.naist.jp} \\
  \And
  Ryo Yonetani\\
  OMRON SINIC X\\
  Tokyo, Japan\\
  \texttt{ryo.yonetani@sinicx.com} \\
}

\begin{document}

\maketitle

\begin{abstract}
\setcounter{footnote}{0}
We present \emph{ShinRL}, an open-source library specialized for the evaluation of reinforcement learning (RL) algorithms from both theoretical and practical perspectives\footnotemark. Existing RL libraries typically allow users to evaluate practical performances of deep RL algorithms through returns. Nevertheless, these libraries are not necessarily useful for analyzing if the algorithms perform as theoretically expected, such as if Q learning really achieves the optimal Q function. In contrast, ShinRL provides an RL environment interface that can compute metrics for delving into the behaviors of RL algorithms, such as the gap between learned and the optimal Q values and state visitation frequencies. In addition, we introduce a flexible solver interface for evaluating both theoretically justified algorithms (\eg, dynamic programming and tabular RL) and practically effective ones (\ie, deep RL, typically with some additional extensions and regularizations) in a consistent fashion. As a case study, we show that how combining these two features of ShinRL makes it easier to analyze the behavior of deep Q learning. Furthermore, we demonstrate that ShinRL can be used to empirically validate recent theoretical findings such as the effect of KL regularization for value iteration \citep{kozunoCVI} and for deep Q learning \citep{vieillard2020munchausen}, and the robustness of entropy-regularized policies to adversarial rewards \citep{husain2021regularized}.
The source code for ShinRL is available on GitHub: \href{https://github.com/omron-sinicx/ShinRL}{https://github.com/omron-sinicx/ShinRL}.
\end{abstract}
\footnotetext{TK devised the main conceptual idea, developed the library, and conducted all the experiments presented in the paper. RY aided in shaping the research and worked in collaboration with TK to write the manuscript.}
\section{Introduction}

Reinforcement learning (RL)~\citep{sutton2018reinforcement} has historically been, and still is, a very active topic in machine learning research. Recent years have particularly seen remarkable progress in research on deep RL, where highly-expressive neural networks are used to approximate policy or Q functions to enable complex sequential decision making. Due to its advantages in dealing with high-dimensional state spaces and learning policies that are generalizable to unseen testing environments, the effectiveness of deep RL has been confirmed in a variety of practical applications, such as robot control~\citep{kober2013reinforcement}, game AI~\citep{mnih2015human}, and economics~\citep{zheng2020ai}, to name a few.

In parallel with research on deep RL for practical tasks, there has been increasing attention paid to efforts to clarify its theoretical basis. Indeed, some state-of-the-art deep RL algorithms can be viewed as an extension of the theoretical foundations of RL such as tabular RL (\ie, no function approximation) and dynamic programming (DP; no exploration while assuming that the complete specification about state transitions is given). Concrete examples of such correspondences between theoretically justified and practically effective algorithms include: from soft Q learning~\citep{haarnoja2017reinforcement} to soft actor-critic (SAC)~\citep{pmlr-v80-haarnoja18b}, from safe policy iteration~\citep{pirotta2013safe} to trust-region policy optimization (TRPO)~\citep{schulman2015trust}, and from conservative value iteration (CVI)~\citep{kozunoCVI} to Munchausen Deep Q learning~\citep{vieillard2020munchausen}. In order to better understand a new deep RL algorithm that has been developed, it is critical to identify its theoretical foundation and validate if it works theoretically as expected, typically under a reasonably simplified setting.

\begin{figure}[t]
\centering
    \includegraphics[width=\linewidth]{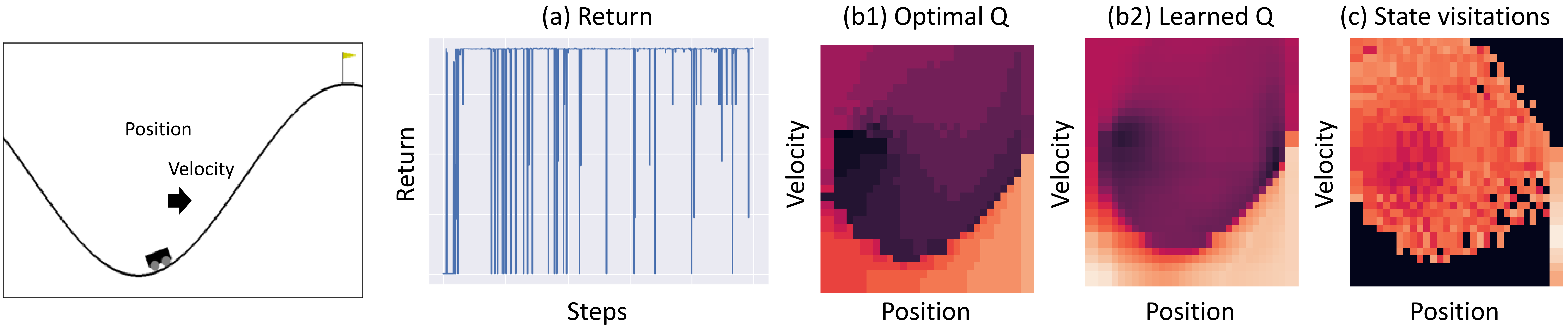}
    \caption{\textbf{Analyzing DQL Results on MountainCar using ShinRL.} (a) Return plot; (b1) Visualization of the optimal Q values, (b2) learned Q values, and (c) state visitation frequency.}
    \label{fig:teaser}
\end{figure}

To this end, one indispensable contribution is the development of open-source libraries that can be used to evaluate RL algorithms from both theoretical and practical perspectives in a principled fashion. Despite many RL libraries have been developed so far~\citep{chainerrl,castro18dopamine,liang2018rllib}, they typically support evaluations of deep RL algorithms only through returns (\ie, cumulative rewards) sampled from episodes. While such evaluations could allow users to systematically compare performances across methods (\eg, \cite{engstrom2020implementation,pmlr-v139-ceron21a}), sampled returns are not necessarily useful for assessing if the methods work as theoretically expected. As a motivating example, suppose a scenario where a user performs a deep Q learning (DQL)~\citep{mnih2015human} on the MountainCar environment~\citep{brockman2016openai}. The user can utilize existing libraries to implement such experiments easily and confirm that the trained network received a high return as shown in \figref{fig:teaser}(a). However, \emph{does this empirical success mean that the network really achieved the optimal Q function?} This is a simple but fundamental question to validate the theoretical expectation that the original (\ie, tabular) Q learning has, which is nonetheless hard to answer for deep RL with non-linear function approximation, even empirically from the returns alone.

Motivated by the observations above, we develop a new RL library named \emph{ShinRL}. At its core, we introduce an RL environment interface that can compute a variety of metrics such as optimal and learned Q functions (\figref{fig:teaser}(b1)(b2)) and state visitation frequency (\figref{fig:teaser}(c)), which are crucial for analyzing the behavior of RL algorithms but are not currently supported in existing libraries. Additionally, ShinRL provides an RL solver interface to evaluate both theoretically-justified and practical RL algorithms in a consistent fashion. By using this interface, users can easily ablate various extensions from developed RL algorithms, such as by removing function approximation and exploration, to empirically evaluate their theoretically-justified variants. ShinRL is implemented JAX\footnote{We also provide another version written in PyTorch in a separate branch.} and can be used without expensive computational resources such as high-end CPUs and GPUs to empirical validate theoretical results that are typically confirmed in reasonably simple environments (eg, \cite{vieillard2020leverage} and \cite{bellemare2016increasing}). Nevertheless, as its main components are built on top of OpenAI Gym~\citep{brockman2016openai}, algorithms once implemented can be immediately available for practical evaluations, such as the ones using Atari~\citep{bellemare2013arcade} with few modifications.

In this paper, we overview the main features of ShinRL and present how they work in practice. Specifically, we first use ShinRL to effectively analyze the behavior of DQL, by clearly visualizing the effects of exploration, function approximation, and more advanced techniques such as double Q learning~\citep{hasselt2010double,van2016deep}. Furthermore, we demonstrate how ShinRL can be used to empirically validate recent theoretical findings in a systematic fashion, such as the effect of KL regularization for value iteration~\citep{kozunoCVI} and for DQL~\citep{vieillard2020munchausen}, and the robustness of entropy-regularized policies to adversarial rewards~\citep{husain2021regularized}.

\section{Background}

\subsection{Preliminaries}
Throughout this paper, we consider an infinite-horizon discounted Markov decision process (MDP) represented by the tuple $\{\mathcal{S},\mathcal{A},P,r,\gamma\}$, where $\mathcal{S}$ is a finite state space, $\mathcal{A}$ is a finite set of actions, $P(s'| s, a)$ is a Markovian transition kernel (where $s, s'\in\mathcal{S}, a\in\mathcal{A}$), $r\in \mathbb{R}^{\mathcal{S}\times \mathcal{A}}$ is a reward function, and $\gamma\in(0,1)$ is a discount factor. The objective of RL is to find the optimal policy $\pi_*$ that maximizes the discounted return (\ie, cumulative reward) given by: 
$\pi_{*} = \argmax_{\pi} \mathbb{E}_{\pi} \left[\sum_{t=0}^{\infty} \gamma^{t} r\left(S_t, A_t\right)\right]$ where $\mathbb{E}_\pi$ is the expectation over all trajectories induced by policy $\pi$.
For a policy $\pi$, the Q function is defined as $Q_\pi(s,a) = \mathbb{E}_\pi\left[\sum_{t=0}^\infty \gamma^t r(S_t,A_t)\middle \vert S_0=s, A_0=a\right]$, the state visitation frequency is defined by $d_{\pi}(s) = (1-\gamma)\sum_{t=0}^{\infty} \gamma^{t} P\left(S_t=s | \pi \right)$. Following \citet{vieillard2020munchausen}, we introduce the component-wise dot product notation $\langle f_1,f_2 \rangle = (\sum_{a} f_1(s,a) f_2(s,a))_{s}\in\mathbb{R}^\mathcal{S}$ for some functions $f_1,f_2\in\mathbb{R}^{\mathcal{S}\times\mathcal{A}}$. With this, the expectation of a Q function over a policy, the V function, can be expressed simply as $V(s) = \langle \pi, Q\rangle (s)=\mathbb{E}_{a\sim \pi(\cdot | s)}[Q(s, a)]$. Further, we introduce another form to describe state transitions: $P v = \left(\sum_{s'} P(s'|s,a) v(s')\right)_{s,a}\in\mathbb{R}^{\mathcal{S}\times\mathcal{A}}$ for $v\in\mathbb{R}^\mathcal{S}$.

\subsection{Dynamic programming, and its extensions to practical RL algorithms}
\label{sec:dp_and_rl}

Classical approaches based on dynamic programming (DP), such as value iteration (VI) and policy iteration (PI), aim to find the optimal Q function as the optimal policy can easily be derived from it. In VI, Q function $Q\in \mathbb{R}^{\mathcal{S}\times \mathcal{A}}$ is iteratively updated by applying a Bellman backup: $Q \leftarrow r + \gamma P \max_a Q$, which is guaranteed to reach the optimality as its unique fixed point is $Q_*$. PI, on the other hand, consists of the following two steps: policy evaluation and policy improvement. The policy evaluation step applies the expected Bellman backup to the Q function: $Q \leftarrow r + \gamma P\langle \pi, Q\rangle$ where its unique fixed point is $Q_\pi$. The policy improvement step updates the policy with the Q function as follows: $\pi \leftarrow \argmax_\pi \langle \pi, Q \rangle$. Alternating these steps leads to the optimal Q function $Q_*$.

Unlike DP-based approaches, RL algorithms typically assume that the state-transition kernel and the reward function are unknown. To achieve the optimal policy or the optimal Q function, they instead require transition samples $(s, a, s', r)$ collected by interacting with the MDP. Notably, many of the RL algorithms are derived from DP. For example, Q-learning~\citep{watkins1992q} can be seen as a variant of VI with exploration, while actor-critic method~\citep{sutton2000policy} is a PI variant with exploration and function approximation of $Q$ and $\pi$.
Many other deep RL algorithms have also been developed by extending DP. Approximate dynamic programming (ADP) is a framework to theoretically analyze RL algorithms using a DP update scheme~\citep{Munos08-finiteTimeAVI,scherrer2015approximate}. Specifically, in the ADP framework, the exploration and function approximation are ``approximated'' as an estimation error, allowing us to analyze how the error propagates to the converged policy.
Doing so has revealed that VI and PI are weak to such errors~\citep{Munos08-finiteTimeAVI,scherrer2015approximate}, which further explains the instability of recent deep Q learning algorithms~\citep{mnih2015human,lillicrap2015continuous,fujimoto2018addressing}.
Some studies have then demonstrated the effectiveness of KL regularization against the error~\citep{azar2012dynamic,ghavamzadeh2011speedy,bellemare2016increasing,vieillard2020leverage,kozunoCVI}, which led to recent KL-regularized deep RL algorithms~\citep{schulman2015trust,vieillard2020momentum,vieillard2020munchausen}. Our main motivation is to develop an open-source library that allows users to reproduce and further explore such connections from theoretical results to practical algorithms.

\section{ShinRL}
\begin{figure}[t]
\centering
    \includegraphics[width=\linewidth]{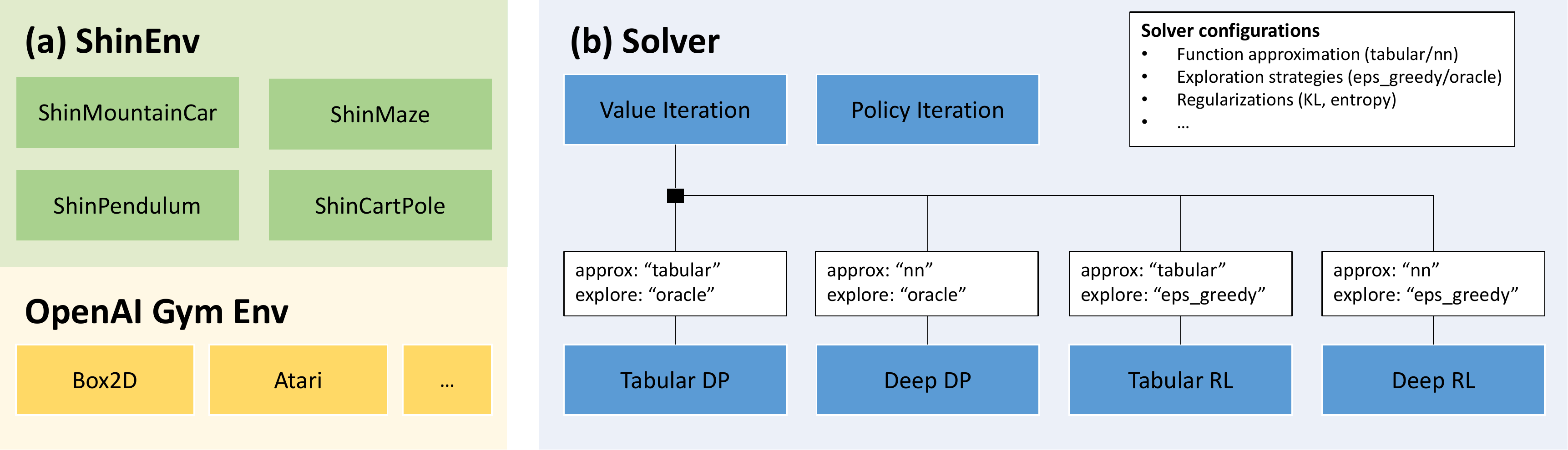}
    \caption{\textbf{Overview of ShinRL.}}
    \label{fig:overview}
\end{figure}

As summarized in \figref{fig:overview}, ShinRL consists of two main modules: \verb+ShinEnv+ as an interface to implement environments modeled by the MDP and \verb+Solver+ as an interface for solving the RL tasks (\ie, finding the optimal policy) on the environments with specified algorithms. In order to maximize the simplicity and flexibility of the library, we keep the number of main modules as low as possible in this way, while also implementing some basic RL necessities such as replay buffers, exploration strategies, and samplers, partially by including external libraries such as \verb+cpprb+~\citep{cpprb}. Using \verb+ShinEnv+ and \verb+Solver+ in combination gives users the ability to evaluate deep RL algorithms as well as their tabular and DP variants through the same interface, making it possible to empirically analyze if developed algorithms work theoretically as expected. In what follows, we describe the design and main features of each module.

\subsection{Environments}
\label{sec:shinenv}
\verb+ShinEnv+ is an interface to implement MDP environments built on top of \verb+Env+ class of OpenAI Gym. We design it to extend Gym's classic control environments with a relatively small state space, such as CartPole and MountainCar, to give users access to the ``oracle'' that can compute exact quantities for returns, Q values, and state visitation frequencies, in an offline fashion. Indeed, when we evaluate a new RL algorithm, we often validate the algorithm on simple and constrained environments before assessing its practical performance under challenging settings (\eg, by using Atari and Mujoco)~\citep{vieillard2020deep,pmlr-v139-ceron21a}. With the existing libraries, we can only collect samples through interactions with environments and only observe estimated returns, typically of high variance, averaged over episodes. On the other hand, exact quantities provided by \verb+ShinEnv+ can help to get more accurate insights into how the algorithm works.

Under the hood of \verb+ShinEnv+, the oracle performs an exhaustive enumeration of all state-action pairs and derives how learned or optimal policies act via sparse matrix calculations. By doing so, \verb+ShinEnv+ provides the following methods: 
\begin{itemize}
\item \verb+calc_q+ computes a Q-value table containing all possible state-action pairs (\ie, $Q_\pi$) given a policy $\pi$. This method accepts some additional input arguments to consider how strongly each reward is affected by KL and entropy regularization imposed on RL algorithms.
\item \verb+calc_optimal_q+ computes the optimal Q-value table (\ie, $Q_*$) by exactly performing value iteration for a specified number of finite-horizon using the precomputed state transition and reward matrices.
\item \verb+calc_visit+ calculates a state visitation frequency table containing all possible states, \ie, $d_\pi$ for a given policy $\pi$.
\item \verb+calc_return+ is a shortcut for computing exact undiscounted returns for a given policy using state transition and reward tables. This is useful as sampling-based approaches just give expected returns typically with high variances.
\end{itemize}

Any environment can be inherited from OpenAI Gym's \verb+Env+ to \verb+ShinEnv+ as long as its state space is reasonably small. When the action space is continuous, we discretize the space with a user-defined number of bins to execute the above-mentioned methods while the environment itself can accept the original continuous actions. Table~\ref{tab:envs} shows some default environments we already implemented. While we developed \verb+ShinCartPole+, \verb+MountainCar+, and \verb+Pendulum+ by inheriting respective OpenAI Gym's \verb+Env+ classes, we create \verb+ShinMaze+ from scratch as an environment that solves an easy 2D maze where agents need to arrive at predefined goal locations while avoiding obstacles, like the one implemented and evaluated in \cite{fu19a}. \rev{For some environments, we also support state spaces given by raw input images, which enforces solvers presented in the next section to automatically use convolutional neural networks when approximating policy or Q functions.}

\begin{table}[t]
\caption{\textbf{Default Environments Implemented in ShinEnv.}}
\label{tab:envs}
\centering
\scalebox{0.9}{
\begin{tabular}{lcccc}
\toprule 
Environment & Discrete action & Continuous action & Image observation & Tuple observation \\
\midrule
\texttt{ShinMaze} & \cmark & \xmark & \xmark & \cmark\\
\texttt{ShinCartPole} & \cmark & \cmark & \xmark & \cmark\\
\texttt{ShinMountainCar} & \cmark & \cmark & \cmark & \cmark\\
\texttt{ShinPendulum} & \cmark & \cmark & \cmark & \cmark\\
\bottomrule
\end{tabular}
}
\end{table}

\begin{figure}[t]
\centering
    \includegraphics[width=\linewidth]{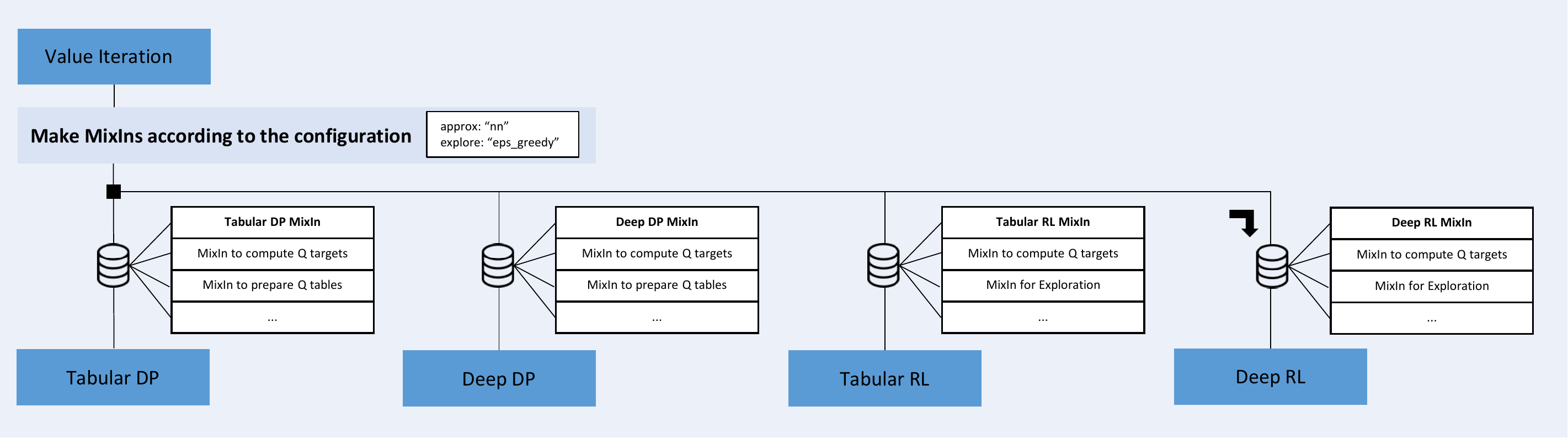}
    \caption{\textbf{Mixin mechanism in ShinRL.}}
    \label{fig:mixin}
\end{figure}

\subsection{Solvers}
\label{sec:solvers}

\verb+Solvers+ is an interface on which a variety of DP and RL algorithms can be implemented. 
As summarized in \figref{fig:overview}(b), \verb+Solvers+ has a hierarchical structure based on how the methods are theoretically related with each other as introduced in \secref{sec:dp_and_rl}. 
More concretely, the current version supports VI and its extensions such as KL-regularized VI (also known as Dynamic Policy Programming~\citep{azar2012dynamic}) and conservative VI~\citep{kozunoCVI}, as well as tabular Q learning, deep Q learning~\citep{mnih2015human}, and Munchausen RL~\citep{vieillard2020munchausen} that are all extended from VI. We also implement PI as well as actor-critic~\citep{konda2000actor} and soft actor-critic~\citep{haarnoja2017reinforcement} as variants of PI.

We design our solver interface to be flexible such that all of these algorithms can be used in a consistent fashion by toggling the solvers' configuration. For example, by disabling the function approximation while enabling exploration, deep RL methods branch to their tabular RL variants. Alternatively, disabling both function approximation and exploration turn the methods back into DP variants. This is very unlike existing libraries that extensively but exclusively support rapid prototyping of deep RL algorithms.

Concretely, we adopt the \verb+mixin+ mechanism to realize the flexible behavior as summarized in \figref{fig:mixin}.
ShinRL's solvers are instantiated by mixing a number of classes called \verb+mixin+, which each define and implement a single feature.
The mixins are chosen according to the solver's configuration and algorithms.
For example, mixins for DQN are chosen by passing \verb+nn+ to \verb+approx+ and \verb+eps_greedy+ to \verb+explore+ in the configuration.
Moreover, this mixin mechanism allows users to easily extend the ShinRL's code base. We will demonstrate its flexibility in the following case studies.
\section{Case Studies}
\label{sec:case_studies}

\begin{figure}[t]
\centering
    \includegraphics[width=\linewidth]{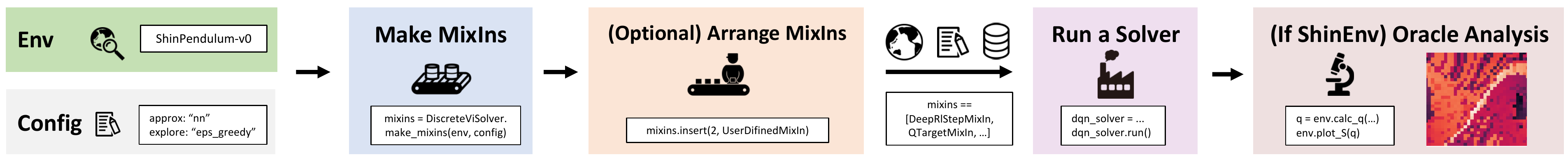}
    \caption{\textbf{Flow of solving a MDP and analyzing the results with ShinRL}}
    \label{fig:flow}
\end{figure}

This section demonstrates how ShinRL can analyze DP and RL algorithms through three case studies. All these studies can be implemented in the same fashion: A solver solves the MDP and then ShinEnv analyzes the results, as summarized in \figref{fig:flow}.

\subsection{Delving into the results of DQL}

\begin{figure}[t]
\centering
    \includegraphics[width=\linewidth]{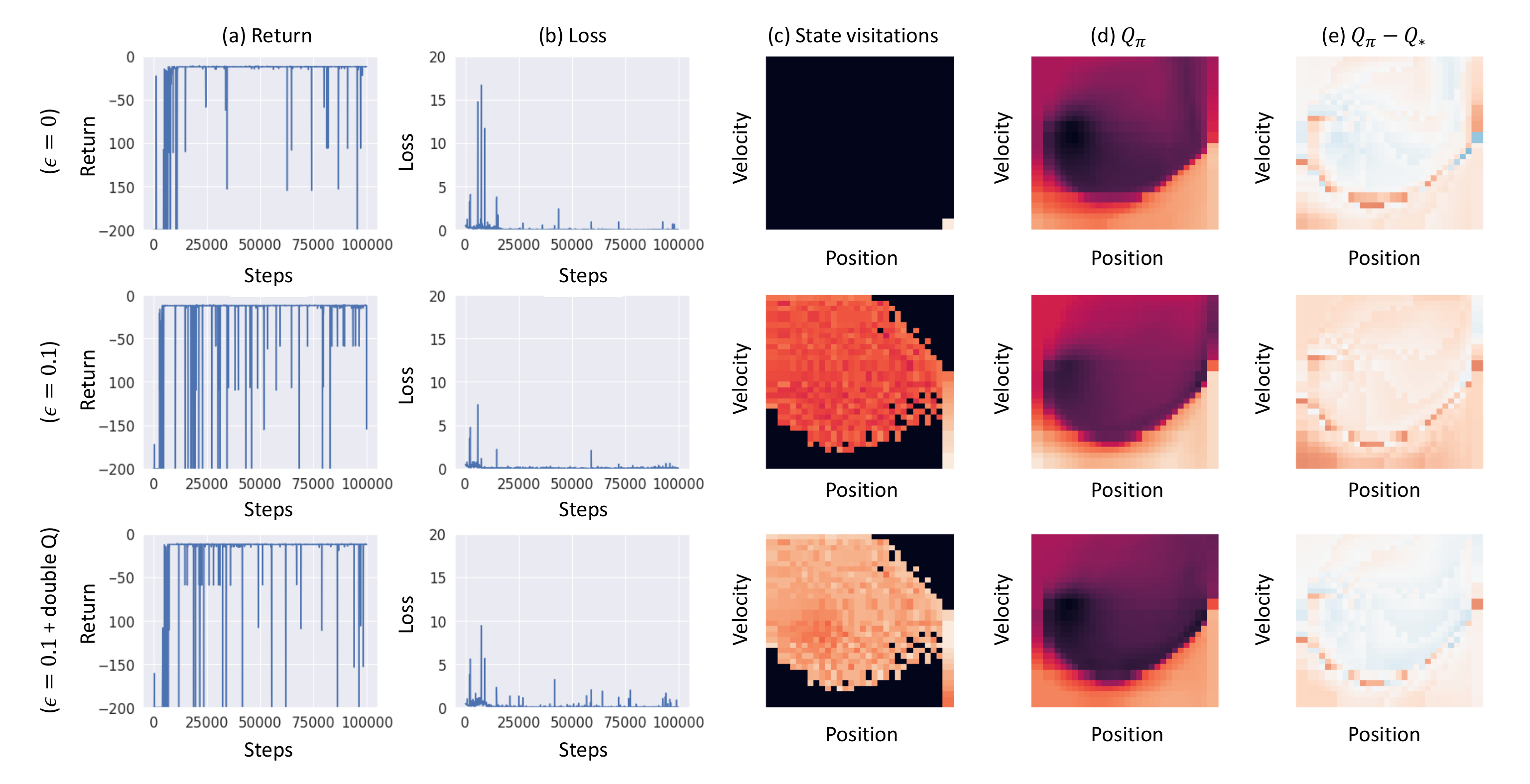}
    \caption{\textbf{Comparison of DQL with Different Settings on ShinMountainCar.} Column (e) shows the gap between the optimal Q values $Q_*$ and learned ones $Q_\pi$, where negative and positive values (\ie, overestimation and underestimation) are highlighted in red and blue.}
    \label{fig:DQL}
\end{figure}

Deep Q learning (DQL)~\citep{mnih2015human} is a popular approach that still has much room for improvement. While deep networks used to approximate the Q function are generally highly expressive, they also need to be trained with diverse transition samples and therefore require a well-tuned exploration strategy in practice.

First, let us introduce how ShinRL can visualize the effectiveness of exploration strategies on the \texttt{ShinMountainCar} environment. As the original DQL can be seen as an extension of Value Iteration with neural network approximation and epsilon-greedy exploration, DQL can be built by passing \verb+nn+ to \verb+approx+ and \verb+eps_greedy+ to \verb+explore+ in the configuration. Instantiating the environment and performing the DQL on it can be done simply in a few lines as shown below.
\begin{lstlisting}[language=python, firstline=1]
import gym
from shinrl import DiscreteViSolver

# instantiate an environment
env = gym.make("ShinMountainCar-v0")

# instantiate deep Q learning-based solver
dql_config = DiscreteViSolver.DefaultConfig(approx="nn", explore="eps_greedy")
dql_mixins = DiscreteViSolver.make_mixins(env, dql_config)
dql_solver = DiscreteViSolver.factory(env, dql_config, dql_mixins)

# run the solver
dql_solver.run()
\end{lstlisting}
The epsilon-greedy exploration strategy highly depends on its value of epsilon, \ie, how likely the agent takes random actions at each step. To understand how this epsilon affects DQL's behaviors, we run two DQL solvers with different constant values set to epsilon, $\epsilon=0.0$ and $\epsilon=0.1$, and observe their state visitation frequencies. This can be done with \verb+calc_visit+ function as follows, which take just about 1ms in a CPU environment\footnote{Confirmed with Intel(R) Core(TM) i7-8700K @ 3.70GHz.}.
\begin{lstlisting}[language=python, firstline=1]
# compute state-action visitation table using learned policy
policy = dql_solver.tb_dict["ExplorePolicy"]
visit = env.calc_visit(policy)
\end{lstlisting}

Figure~\ref{fig:DQL}(a) and (b) present plots for returns and losses, and the corresponding state visitation tables are visualized in (c). A bit surprisingly, both solvers finally solved the task (defined by the return arrived at $-20$), and their losses are almost comparable. Nevertheless, they demonstrate a clear difference in state visitation frequencies, where the solver with $\epsilon=0$ leads to a poor exploration policy that can potentially visit a limited set of states even after a large number of steps, while the solver with $\epsilon=0.1$ can visit almost all possible states that the agent can reach.
Now we are interested in how this difference in state visitation frequencies affects the quality of learned Q functions. Here, we visualize the difference between learned and optimal Q values as follows:
\begin{lstlisting}[language=python, firstline=1]
optimal_q = env.calc_optimal_q()
learned_q = dql_solver.tb_dict["Q"]
diff_q = learned_q - optimal_q
\end{lstlisting}
\rev{As shown in \figref{fig:DQL}(e), Q values are inaccurate in many places due to underestimation under $\epsilon=0$ and overestimation under $\epsilon=0.1$. Overestimation is particularly a known phenomenon and can be alleviated via double-Q learning~\citep{hasselt2010double,van2016deep} that learns two Q functions with different sets of samples. The bottom row of \figref{fig:DQL} shows that the double-Q trick indeed improves the accuracy of the learned Q values, except for some state-action pairs that were not visited during learning. This further implies the importance of better dealing with out-of-distribution actions such as done in offline RL~\citep{levine2020offline}, which we leave for future work.}

\subsection{Comparing VI, KL-regularized VI, CVI, and Munchausen DQL}

\begin{figure}[t]
\centering
    \includegraphics[width=\linewidth]{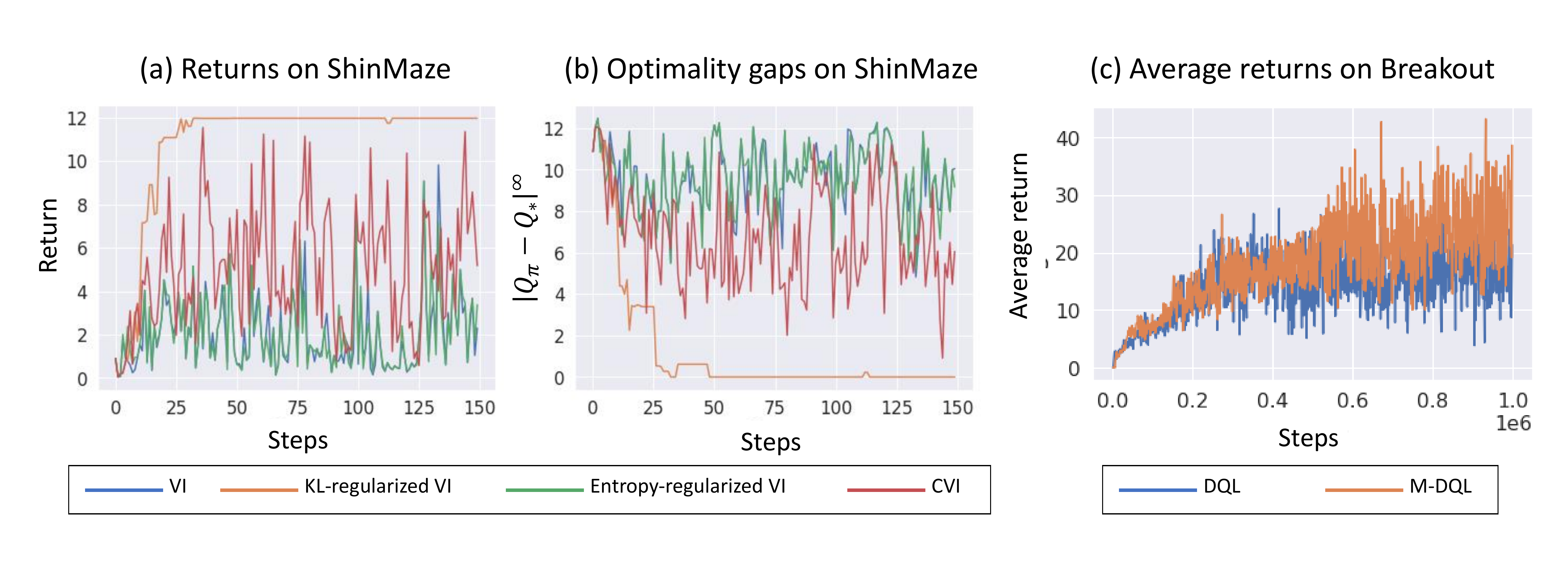}
    \caption{\textbf{Comparisons of VI Variants with Various Regularizations.} (a) Return plots and (b) optimality gaps given by $\|Q_{\pi_k} - Q_*\|_\infty$ for VI, KL-regularized VI, entropy-regularized VI, and CVI on the ShinMaze environment. (c) DQL and M-DQL on the Breakout environment.}
    \label{fig:CVI}
\end{figure}

As we introduced in \secref{sec:dp_and_rl}, VI theoretically becomes robust to \rev{estimation errors, which typically arise due to function approximation and exploration,} by imposing KL regularization~\citep{vieillard2020leverage}. Furthermore, involving entropy as well as KL regularizations turns VI to conservative VI (CVI)~\citep{kozunoCVI}, which inspires the formulation of Munchausen DQL (M-DQL)~\citep{vieillard2020munchausen} that extends conventional deep Q learning with these regularizations to improve the stability. In this case study, \rev{we demonstrate how these theoretical findings of VI, KL-regularized VI, CVI, DQL, and M-DQL can be confirmed empirically and systematically using ShinRL.}

To observe how VI changes its behavior with regularizations, let us first introduce the following DP update scheme:
\begin{equation}\label{eq:kl-er-dp}
    \begin{cases}
        \pi_{k+1} &= \text{argmax}_{\pi} \left(\langle \pi, Q_k\rangle -\tau \text{KL}(\pi \| \pi_k) + \lambda \mathcal{H}(\pi)\right), \\
        Q_{k+1} &= r + \gamma P \langle \pi_{k+1}, Q_k - \tau \text{KL}(\pi_{k+1} \| \pi_k) + \lambda \mathcal{H}(\pi_{k+1}) \rangle + \epsilon_k,
    \end{cases}
\end{equation}
where $\tau$ and $\lambda$ are coefficients for KL and entropy regularization, respectively. $\epsilon_k\sim \mathcal{N}(0, \sigma)$ is a zero-mean Gaussian error vector with standard deviation $\sigma$ at $k$-th iteration, which models either function approximation and/or exploration errors. 
Without the error term $\epsilon_k$, evaluating this regularized DP with ShinRL is quite simple by just toggling the configurations of \verb+ViSolver+, where the parameters $\tau$, $\lambda$ are respectively specified by \verb+kl_coef+, \verb+er_coef+. 
For example, mixins for CVI that comes with both KL and entropy regularizations can be instantiated as follows:

\begin{lstlisting}[language=python, firstline=1]
# Instantiate VI
cvi_config = DiscreteViSolver.DefaultConfig(
    approx="tabular",  # use tabular method
    explore="oracle",  # use oracle for exploration
    kl_coef=0.1,  
    er_coef=0.3,
)
cvi_mixins = DiscreteViSolver.make_mixins(env, cvi_config)
\end{lstlisting}

The error term can be easily implemented by arranging the mixins.
We implement \eqref{eq:kl-er-dp} by adding a mixin which add noise to computed Q values:
\begin{lstlisting}[language=python, firstline=1]
class ErrorMixIn:
    # A mixin to add noise to the computed Q values
    ...

mixins = [ErrorMixIn, *cvi_mixins]  # Add ErrorMixIn to CVI's mixins
cvi_solver = DiscreteViSolver.factory(env, cvi_config, mixins)
cvi_solver.run()
\end{lstlisting}

Figure \ref{fig:CVI} shows (a) returns as well as (b) the optimality gap defined by $\|Q_*-Q_\pi\|_\infty$ on the \verb+ShinMaze+ environment. KL-regularized VI is indeed robust against noise empirically and can easily reach the optimality. On the other hand, entropy regularization is less important than KL regularization, which is also explained by \citet{vieillard2020leverage}. Nevertheless, the next section will show the effectiveness of entropy regularization when used in the SAC algorithm~\citep{pmlr-v139-ceron21a}.

Now the task is to compare DQL and M-DQL. They are deep RL algorithms that should be evaluated in more challenging environments to best show their performances. To this end, ShinRL fully supports OpenAI Gym, and can call the \verb+minatar+ environment~\citep{young2019minatar} that is a lightweight testbed inspired by Atari games.
\begin{lstlisting}[language=python, firstline=1]
from shinrl import make_minatar
env = make_minatar("breakout")

# M-DQL
mdql_config = DiscreteViSolver.DefaultConfig(
    "approx": "nn",
    "explore": "eps_greedy",
    # the solver reduces to standard DQL by setting kl_coef and er_coef to zero.
    kl_coef=0.027,
    er_coef=0.003
)
mdql_mixins = DiscreteViSolver.make_mixins(env, mdql_config)
mdql_solver = DiscreteViSolver.factory(env, mdql_config, mdql_mixins)
mdql_solver.run()
\end{lstlisting}
Figure~\ref{fig:CVI} (c) depicts return plots for the Breakout environment, demonstrating that M-DQL outperforms DQL thanks to KL and entropy regularizations.

\subsection{Evaluating robustness of the SAC algorithm to adversarial rewards}

Another family of algorithms that ShinRL supports extensively is policy iteration (PI), which is the foundation of many recent deep RL algorithms. For example, the SAC algorithm~\citep{pmlr-v80-haarnoja18b} is an extension of PI with function approximation, exploration, and entropy regularization. Some recent work shows that the SAC algorithm can learn a robust policy, both empirically~\citep{pmlr-v80-haarnoja18b} and theoretically~\citep{husain2021regularized}, thanks to its entropy regularizer. In this case study, we first investigate how SAC changes its robustness, in particular to adversarial rewards presented by \cite{husain2021regularized}, with different entropy regularization coefficient $\lambda$ in the \verb+ShinPendulum+ environment. In ShinRL, mixins for SAC can be instantiated by \verb+PiSolver+ as follows:
\begin{lstlisting}[language=python, firstline=1]
# Instantiate SAC
sac_config = DiscretePiSolver.DefaultConfig(
    approx="nn",  # use neural network approximation
    explore="eps_greedy",  # use epsilon-greedy policy for exploration
    er_coef=0.1,
)
sac_mixins = DiscretePiSolver.make_mixins(env, sac_config)
\end{lstlisting}
Note that by setting \verb+er_coef+ to $0$, the method reduces to the vanilla actor-critic algorithm~\citep{konda2000actor}.
As done in the experiments of \citet{husain2021regularized}, we consider the following adversarial reward $r_\text{adv}$ by slightly modifying the original implementation of the pendulum: 
\begin{equation}\label{eq:adv-rew} 
    r_\text{adv} = 
    \begin{cases}
        r(s, a) + \epsilon \; &\text{if}\; r(s, a) \leq -5 \\
        r(s, a) \; &\text{otherwise},
    \end{cases}
\end{equation}
where $\epsilon$ is sampled from the normal distribution $\mathcal{N}(4.9, 0.1)$.
This reward design promotes the agent to stay the pendulum around its initial state while the optimal behavior is still swinging it up.
Similar to the previous case study, we implemented \eqref{eq:adv-rew} on ShinRL's SAC by adding another mixin which assigns adversarial rewards to the collected data.

\begin{figure}[t]
\centering
    \includegraphics[width=\linewidth]{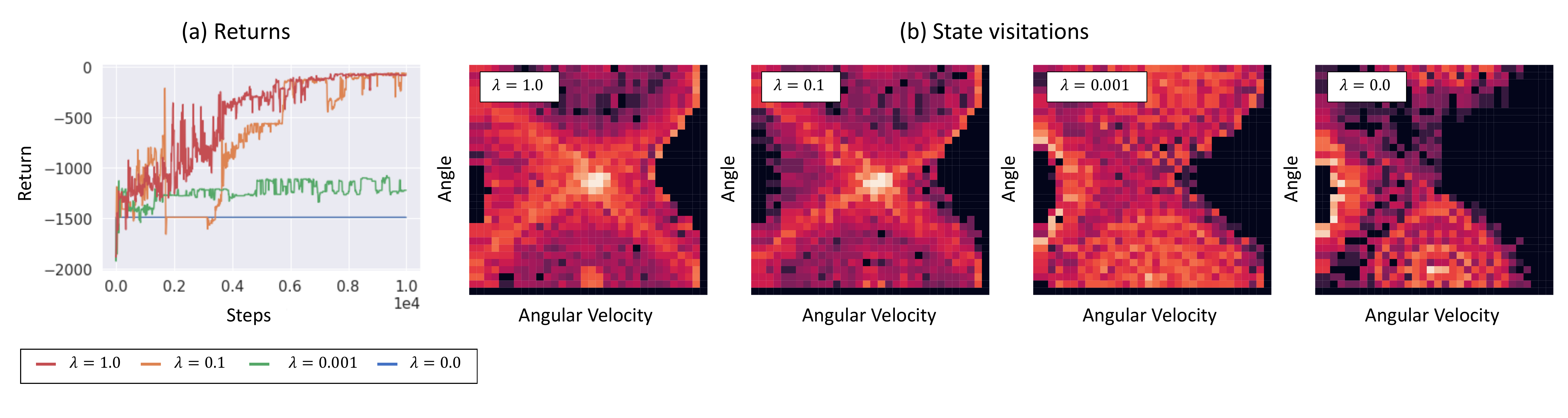}
    \caption{\textbf{SAC with Different Coefficients $\lambda$ for Entropy Regularization:} (a) Return plots and (b) State visitation frequencies (higher frequencies highlighted in red) on the ShinPendulum environment.}
    \label{fig:SAC}
\end{figure}

Figure \ref{fig:SAC}(a) shows the learning curves of SAC with different coefficients for entropy regularization. We confirm that reasonably increasing the regularization strength improves the robustness against adversarial rewards as confirmed by \citet{husain2021regularized}.

Furthermore, we validate the finding of \citet{haarnoja2017reinforcement} that empirically confirms the improvement of exploration quality as the entropy regularization becomes stronger. By using \verb+count_visit+ function, we can visualize the frequencies of state-action pairs stored in a replay buffer, making it possible to assess if the exploration is sufficient during the training. As shown in \figref{fig:SAC}(b), we confirm that a wider range of state-action pairs are visited as $\lambda$ becomes higher.

\section{Conclusion}
We presented ShinRL, an open-source library that can evaluate RL algorithms from both theoretical and practical perspectives in a principled fashion. As shown in our case studies, ShinRL can be used to analyze the behavior of deep RL algorithms through the lens of Q-value tables and state visitation frequencies, which are not immediately available in existing RL libraries. Further, we empirically confirm recent theoretical findings of KL regularization and entropy regularization for RL~\citep{kozunoCVI,vieillard2020munchausen,husain2021regularized} using our flexible RL solver interface.
Future work will seek to extend the library to deal with a wider variety of tasks and algorithms not only on RL but also imitation learning and offline RL.

\section*{Acknowledgments}
The authors would like to thank Masashi Hamaya for helpful feedback on the manuscript.

\bibliographystyle{unsrtnat}
\bibliography{reference}

\end{document}